\documentclass[letterpaper, 10 pt, journal, twoside]{IEEEtran}
\IEEEoverridecommandlockouts                              % This command is only needed if 
\usepackage{hyperref} 
\usepackage{multicol}
\usepackage{float}
\usepackage[caption = false]{subfig}
\usepackage[final]{graphicx}
\usepackage{listings}
\usepackage[style=ieee]{biblatex}
\usepackage{xcolor}
\usepackage{pgfplots}
\usepgfplotslibrary{statistics}
\usepackage[export]{adjustbox}

\newcommand{\rebuttal}[1]{#1}

\title{
The Hilti SLAM Challenge Dataset
}

\author{Michael Helmberger$^{1}$, Kristian Morin$^{1}$, Beda Berner$^{1}$, Nitish Kumar$^{1}$, \\Giovanni Cioffi$^{2}$, Davide Scaramuzza$^{2}$% <-this % stops a space
\thanks{Manuscript received: February, 24, 2022; Revised May, 17, 2022; Accepted May, 30, 2022.}

\thanks{This paper was recommended for publication by Editor Javier Civera upon evaluation of the Associate Editor and Reviewers' comments.

This work was supported by the Hilti AG, Schaan, Liechtenstein.}

\thanks{$^{1}$ Hilti AG, Schaan, Liechtenstein. $^{2}$ Robotics and Perception Group, Department of Informatics, University of Zurich, and Department of Neuroinformatics, University of Zurich and ETH Zurich, Switzerland (\protect\url{http://rpg.ifi.uzh.ch}). {\tt\footnotesize cioffi@ifi.uzh.ch}}%
\thanks{Digital Object Identifier (DOI): see top of this page.}
}

\begin{document}

\maketitle
%\thispagestyle{empty}
%\pagestyle{empty}

%%%%%%%%%%%%%%%%%%%%%%%%%%%%%%%%%%%%%%%%%%%%%%%%%%%%%%%%%%%%%%%%%%%%%%%%%%%%%%%%
\begin{abstract}

Research in Simultaneous Localization and Mapping (SLAM) has made outstanding progress over the past years.
SLAM systems are nowadays transitioning from academic to real world applications.
However, this transition has posed new demanding challenges in terms of accuracy and robustness.
To develop new SLAM systems that can address these challenges, new datasets containing cutting-edge hardware and realistic scenarios are required.
We propose the \textit{Hilti SLAM Challenge Dataset}.
Our dataset contains indoor sequences of offices, labs, and construction environments and outdoor sequences of construction sites and parking areas.
All these sequences are characterized by featureless areas and varying illumination conditions that are typical in real-world scenarios and pose great challenges to SLAM algorithms that have been developed in confined lab environments.
%Accurate ground truth, at millimeter level, is provided for each sequence.
\rebuttal{Accurate sparse ground truth, at millimeter level, is provided for each sequence.}
The sensor platform used to record the data includes a number of visual, lidar, and inertial sensors, which are spatially and temporally calibrated.
The purpose of this dataset is to foster the research in sensor fusion to develop SLAM algorithms that can be deployed in tasks where high accuracy and robustness are required, \textit{e.g.}, in construction environments.
Many academic and industrial groups tested their SLAM systems on the proposed dataset in the \textit{Hilti SLAM Challenge}.
The results of the challenge, which are summarized in this paper, show that the proposed dataset is an important asset in the development of new SLAM algorithms that are ready to be deployed in the real-world.
\end{abstract}


%-------------------------------------------------------------------------
% SECTIONS
\section*{Supplementary Material} \label{sec:SupplementaryMaterial}
The dataset as well as information regarding the \textit{Hilti SLAM Challenge} is available at \url{https://www.hilti-challenge.com}. 
% At the bottom of the page there is a leader board with the ranking of the systems submitted and evaluated in the challenge so far.
The results of the \textit{Hilti SLAM Challenge} were also presented in a talk: \url{https://www.youtube.com/watch?v=3oqTGrnSkrY&t=685s}
\section{Introduction} \label{sec:Introduction}

\IEEEPARstart{R}{ecent} years have seen outstanding progresses in SLAM~\cite{Cadena16tro}.
The transition from demonstrative to real-world applications is happening at this moment.
A promising application of SLAM for autonomous robotics is in construction sites. 
Construction robotics offers a way to remove hazards for workers, improve task productivity, and collect high-quality data~\cite{jaibot}.
However, these environments bring many challenges to SLAM systems. Featureless scenes, varying illumination conditions, and sudden motions are among the main challenges.

The deployment of SLAM algorithms to real-world applications has shed light on the limitations of the current systems.
These limitations are being addressed by academic research often in collaboration with industrial partners~\cite{SLAMCORE}.
We believe that collaborations between academia and industry have the potential to accelerate the process of developing new SLAM systems that are able to meet tight requirements in terms of accuracy and robustness.

An important role in this phase is played by the availability of datasets containing relevant scenarios and sensors.
The scenarios should portray actual real-world applications where SLAM systems are deployed.
As the robotic community has shown in several works~\cite{Leutenegger15ijrr, cioffi2022continuous, debeunne2020review}, the highest accuracy and robustness is achieved by fusing multiple and complementary sensors.
For this reason, SLAM datasets should contain multiple sensor modalities.

\begin{figure}[t]
  \centering
  \includegraphics[width=\columnwidth]{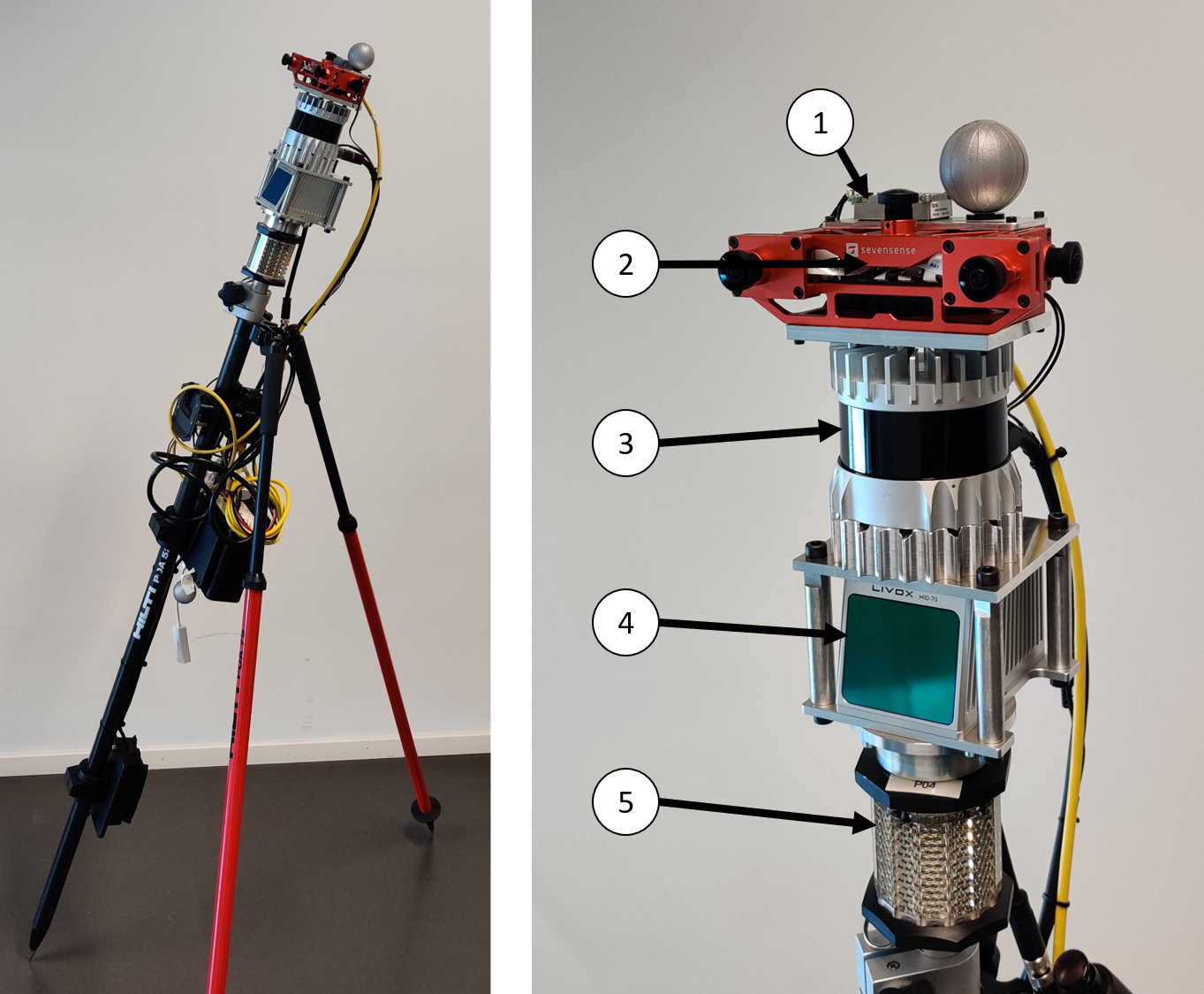}
  \caption{\textit{Left}: The full sensor stick resting on a tripod. \textit{Right}: (1) ADIS16445 IMU (2) AlphaSense 5-camera module (3) Ouster OS0-64 LiDAR (4) Livox MID70 LiDAR (5) prism for the Total Station. }
  \label{figurelabel_phasma_stick}
\end{figure}

\begin{figure*}[!htp]
\subfloat[Basement]{\includegraphics[width = 4.1cm]{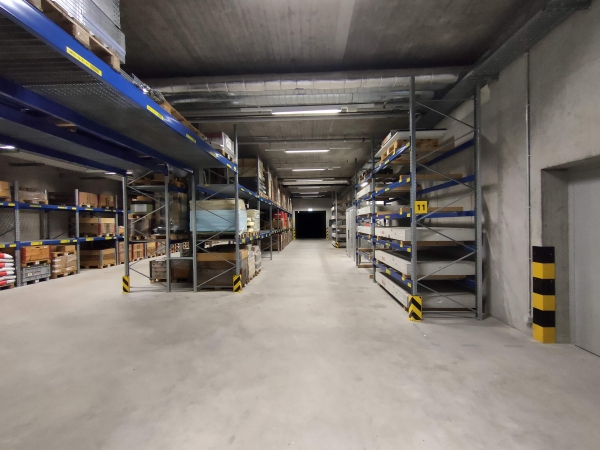}} 
\hspace{1em}% Space between image A and B
\subfloat[Campus]{\includegraphics[width = 4.1cm]{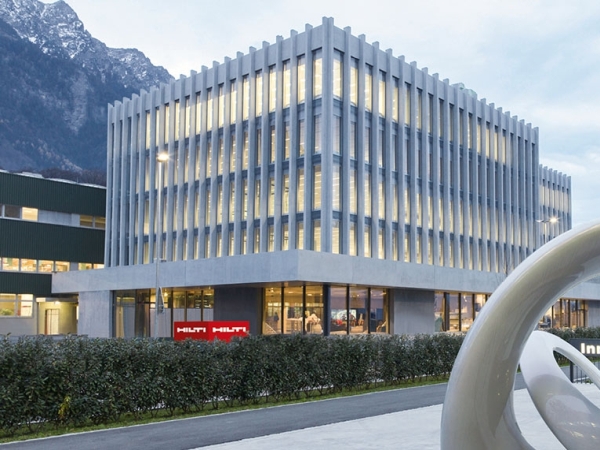}}
\hspace{1em}% Space between image A and B
\subfloat[Construction Site]{\includegraphics[width = 4.1cm]{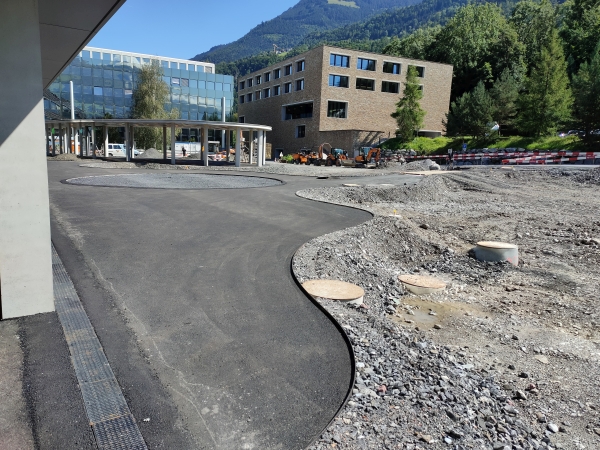}}
\hspace{1em}% Space between image A and B
\subfloat[IC Office]{\includegraphics[width = 4.1cm]{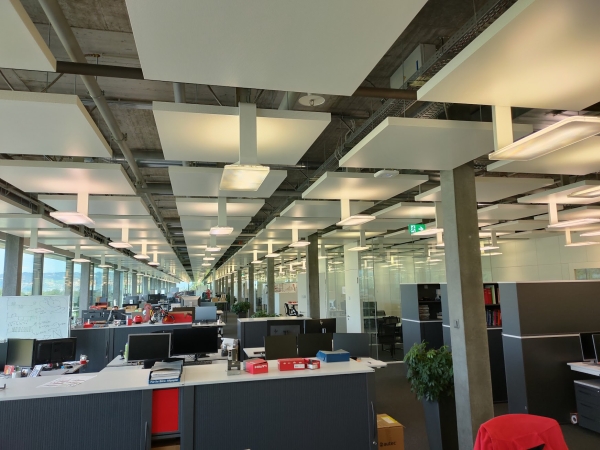}} \\
\hspace{1em}% Space between image A and B
\subfloat[Lab]{\includegraphics[width = 4.1cm]{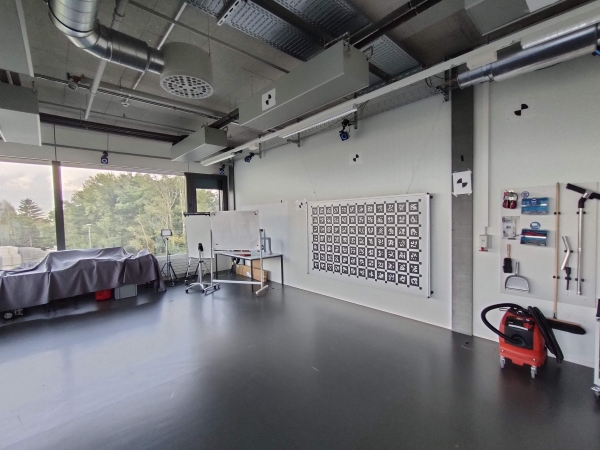}} 
\hspace{1em}% Space between image A and B
\subfloat[Office Mitte]{\includegraphics[width = 4.1cm]{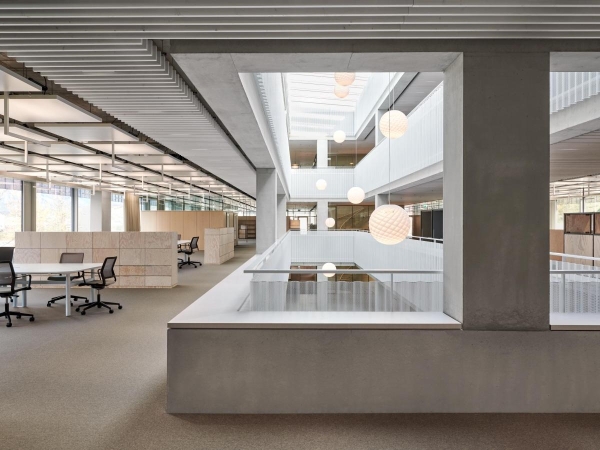}} 
\hspace{1em}% Space between image A and B
\subfloat[Parking]{\includegraphics[width = 4.1cm]{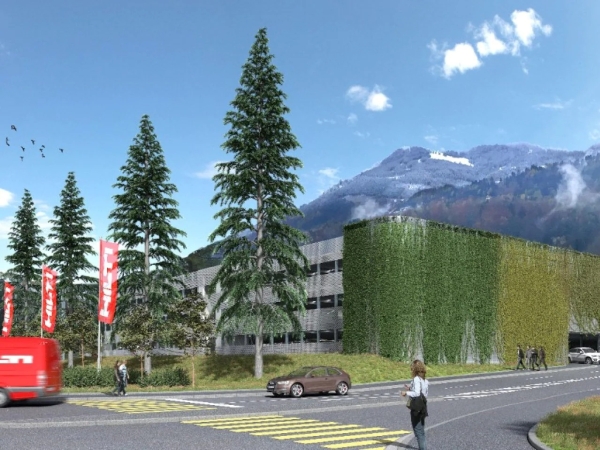}} 
\hspace{1em}% Space between image A and B
\subfloat[RPG Tracking Area]{\includegraphics[width = 4.1cm]{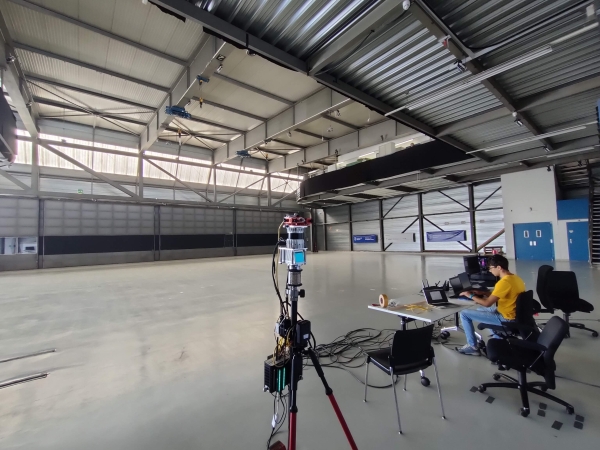}} 
\caption{Locations where the dataset have been captured.}
\label{some_example_scenes}

\end{figure*}

Many SLAM datasets have been proposed over the past years~\cite{Burri25012016, Geiger2012CVPR, blanco2014malaga, ramezani2020newer, zhang2021multi, pfrommer2017penncosyvio, tartanair2020iros, schubert2018tum, zuniga2020vi, carlevaris2016university, Delmerico19icra, Mueggler_2017}, each of them has specific contributions on the availability of multiple sensors and the type of scenarios and motions.
Most of the datasets~\cite{Burri25012016, blanco2014malaga, pfrommer2017penncosyvio, tartanair2020iros, schubert2018tum, zuniga2020vi, carlevaris2016university, Delmerico19icra} focus on visual and inertial data, while only a few~\cite{Geiger2012CVPR, ramezani2020newer, zhang2021multi} also provide LiDAR data.
Depending on the type of environment and motion, each dataset poses different challenges to SLAM.
Visual and inertial data recorded onboard an unmanned aerial vehicle (UAV) in~\cite{Burri25012016, Delmerico19icra} bring difficulties to visual-inertial odometry (VIO) and SLAM systems because of fast motions.
Varying illumination conditions and moving objects are the main challenges of datasets recorded in autonomous driving scenarios~\cite{blanco2014malaga,Geiger2012CVPR, janai2020computer}.
%\davide{Please add the book by ANdreas Geiger on Vision for Autonomous Driving: \url{https://www.nowpublishers.com/article/Details/CGV-079}}
Sudden motions and low texture, as well as dynamic illumination conditions, are the main challenges of datasets recorded with a handheld sensor platform~\cite{ramezani2020newer, zhang2021multi, pfrommer2017penncosyvio, schubert2018tum, zuniga2020vi}.

We propose the \textit{Hilti SLAM Challenge Dataset}.
The purpose of this dataset is to foster the research in sensor fusion to develop SLAM algorithms that can be deployed in tasks where high accuracy, at millimeter level, and robustness are required, \textit{e.g.}, in construction environments.
Our dataset contains indoor sequences of construction sites, offices, and labs and outdoor sequences of construction environments and parking areas (in Fig.~ \ref{some_example_scenes}).
All these sequences are characterized by featureless areas and varying illumination conditions that are typical of real-world scenarios and pose challenges to SLAM algorithms that have been developed in confined lab environments.
Millimeter accurate ground truth by mean of a motion capture system (MoCap) or a Total Station~\cite{8592768}, is provided for each sequence.
To record the data, we have created a suite of sensors including a number of visual, LiDAR, and inertial sensors, using the latest products in commercially available sensing technologies with particular attention made to time synchronization and spatial calibration.
With the use of redundant sensors, this dataset also directly compares sensor performance in different environments, which can be informative for the design of SLAM systems.
With this dataset, we aim to stimulate research on robust indoor positioning, mapping, and navigation with the particular application to construction environments.

This dataset was used in the \textit{Hilti SLAM Challenge}, where 
many academic and industrial groups submitted their SLAM systems.
The results of the challenge, which are summarized in Section~\ref{sec:Challenge}, show that the proposed dataset can be an important asset in the development of new SLAM algorithms that are ready to be deployed in the real-world.

To summarize, the main contributions of the \textit{Hilti SLAM Challenge Dataset} are:
\begin{itemize}
    \item Real-world sequences recorded in indoor offices and labs, indoor and outdoor construction environments, and outdoor parking areas containing challenging featureless areas and varying illumination conditions.
    \item Sensor suite containing 5 cameras (1 stereo pair), 2 lidars, and 3 IMUs with accurate spatial and temporal calibration.
    \item The \textit{Hilti SLAM Challenge}. Many academic and industrial groups used this dataset to further develop their SLAM systems and take part in the \textit{Hilti SLAM Challenge} competition.
\end{itemize}
\section{Related Work} \label{sec:RelatedWork}

\begin{table*}[t]
\vspace*{0.5cm} 
\caption{Comparison of SLAM datasets}
\label{tab:datasets}
\begin{tabular}{ccccccc}
\hline
\textbf{Dataset} &
  \textbf{Num. seq,} &
  \textbf{Environment} &
  \textbf{Motion} &
  \textbf{Sensors} &
  \textbf{Synchronization} &
  \textbf{Ground truth} \\ \hline
EuRoC~\cite{Burri25012016} &
  11 &
  Indoor &
  UAV &
  \begin{tabular}[c]{@{}c@{}}1 Stereo camera\\ 1 IMU\end{tabular} &
  Hw &
  \begin{tabular}[c]{@{}c@{}}Laser tracker,\\ MoCap\end{tabular} \\ \hline
KITTI~\cite{Geiger2012CVPR} &
  22 &
  Outdoor &
  Car &
  \begin{tabular}[c]{@{}c@{}}1 Stereo camera\\ 1 Stereo RGB camera\\ 1 Lidar\\ 1 IMU\end{tabular} &
  Sw &
  GNSS-INS \\ \hline
Malaga~\cite{blanco2014malaga} &
  15 &
  Outdoor &
  Car &
  \begin{tabular}[c]{@{}c@{}}1 Stereo RGB camera\\ 1 IMU\end{tabular} &
  Sw &
  GPS \\ \hline
\begin{tabular}[c]{@{}c@{}}The Newer \\ College Dataset~\cite{zhang2021multi}\end{tabular} &
  3 &
  In- / Outdoor &
  Handheld &
  \begin{tabular}[c]{@{}c@{}}4 Cameras\\ 1 Lidars\\ 2 Imus\end{tabular} &
  Hw / Sw &
  \begin{tabular}[c]{@{}c@{}}3D imaging laser \\ scanner, ICP\end{tabular} \\ \hline
PennCOSYVIO~\cite{pfrommer2017penncosyvio} &
  4 &
  In- / Outdoor &
  Handheld &
  \begin{tabular}[c]{@{}c@{}}4 RGB cameras\\ 1 Stereo camera\\ 1 Fisheye camera\\ 2 IMUs\end{tabular} &
  Hw / Sw &
  \begin{tabular}[c]{@{}c@{}}Fiducial \\ markers\end{tabular} \\ \hline
TartanAir~\cite{tartanair2020iros} &
  30 &
  In- / Outdoor &
  Simulation &
  \begin{tabular}[c]{@{}c@{}}Synthetic \\ 1 stereo RGB camera\end{tabular} &
  Sim. &
  Simulation \\ \hline
TUM VIO~\cite{schubert2018tum} &
  28 &
  In- / Outdoor &
  Handheld &
  \begin{tabular}[c]{@{}c@{}}1 Stereo camera\\ 1 IMU\end{tabular} &
  Hw &
  \begin{tabular}[c]{@{}c@{}}MoCap \\ (start, end)\end{tabular} \\ \hline
UMA VI~\cite{zuniga2020vi} &
  32 &
  In- / Outdoor &
  Handheld &
  \begin{tabular}[c]{@{}c@{}}1 Stereo RGB camera\\ 1 Stereo camera\\ 1 IMU\end{tabular} &
  Hw &
  \begin{tabular}[c]{@{}c@{}}Pose alignment\\ (start, end)\end{tabular} \\ \hline
UMich~\cite{carlevaris2016university} &
  27 &
  In- / Outdoor &
  UGV &
  \begin{tabular}[c]{@{}c@{}}6 RGB cameras\\ 1 IMU\end{tabular} &
  Sw &
  \begin{tabular}[c]{@{}c@{}}Fusion of GPS, \\ IMU, laser\end{tabular} \\ \hline
UZH-FPV~\cite{Delmerico19icra} &
  28 &
  In- / Outdoor &
  UAV &
  \begin{tabular}[c]{@{}c@{}}1 Stereo camera\\ 1 Event camera\\ 2 IMUs\end{tabular} &
  Hw / Sw &
  \begin{tabular}[c]{@{}c@{}}Fusion of vision,\\ IMU, laser\end{tabular} \\ \hline
\begin{tabular}[c]{@{}c@{}}The Hilti SLAM \\ Challenge Dataset (\textbf{Ours})\end{tabular} &
  12 &
  In- / Outdoor &
  Handheld &
  \begin{tabular}[c]{@{}c@{}}5 Cameras\\ 2 Lidars\\ 3 Imus\end{tabular} &
  Hw / Sw &
  \begin{tabular}[c]{@{}c@{}}MoCap, \\ Total Station\end{tabular} \\ \hline
\end{tabular}
\end{table*}

Among the plethora of datasets proposed to benchmark SLAM, as well as VIO, systems,~\cite{Burri25012016, Geiger2012CVPR, blanco2014malaga, ramezani2020newer, zhang2021multi, pfrommer2017penncosyvio, tartanair2020iros, schubert2018tum, zuniga2020vi, carlevaris2016university, Delmerico19icra, Mueggler_2017} are the ones that mostly relate to the \textit{Hilti SLAM Challenge Dataset}.
Many datasets~\cite{Burri25012016, blanco2014malaga, pfrommer2017penncosyvio, tartanair2020iros, schubert2018tum, zuniga2020vi, carlevaris2016university, Delmerico19icra} focus on visual and inertial data.
The datasets~\cite{Burri25012016,Delmerico19icra} provide hardware-synchronized visual, from a stereo camera, and inertial data recorded on board an UAV.
They challenge SLAM algorithms because of fast motions, with~\cite{Delmerico19icra} containing the most aggressive maneuvers.
For both datasets, the ground truth is generated by solving a bundle adjustment problem including visual, inertial, and laser tracker data.
In~\cite{Burri25012016}, some sequences provide ground truth data for a MoCap system.
The datasets~\cite{pfrommer2017penncosyvio, schubert2018tum, zuniga2020vi} provide visual and inertial data recorded using a handheld sensor platform.
In all these datasets, the sensor platform includes a stereo camera and an IMU.
In addition, stereo RGB cameras are also available in~\cite{pfrommer2017penncosyvio,zuniga2020vi}.
The ground truth in~\cite{pfrommer2017penncosyvio} is obtained by localizing to fiducial markers, which are placed along the trajectory.
The ground truth at the start and end of the trajectory is obtained by pose alignment with respect to fiducial markers and using a MoCap system in~\cite{zuniga2020vi} and~\cite{schubert2018tum}, respectively.
In~\cite{tartanair2020iros}, it was proposed a large (30 sequences) simulated dataset containing a broad range of scenarios.
The dataset in~\cite{Mueggler_2017} includes data from an event camera, a survey on event vision is presented in ~\cite{gallego2020event}, as well as a standard camera and IMU.
It was designed to incentivize researchers to investigate the use of event cameras in SLAM.
The datasets~\cite{blanco2014malaga,Geiger2012CVPR,carlevaris2016university} contain data recorded in autonomous driving, and ground robot navigation, scenarios.
They all include RGB cameras and an IMU.
The dataset~\cite{Geiger2012CVPR} is one of the most used datasets for benchmarking SLAM systems in autonomous driving.
It also includes LiDAR data.
The ground truth for these datasets is provided by inertial navigation systems (INS) aided by global navigation satellite systems (GNSS).
The dataset proposed in~\cite{zhang2021multi}, which is an extension of~\cite{ramezani2020newer}, is the closest to our dataset.
It contains sequences recorded in a university campus.
The sensor platform includes 4 cameras, 1 LiDAR, and 2 IMUs.
The ground truth is provided by an iterative closest point~\cite{besl1992method} (ICP) algorithm that registers the LiDAR point cloud to a prior map.
Differently from~\cite{zhang2021multi}, our dataset provides millimeter accurate ground truth from the Hilti PLT 300~\footnote{\url{https://www.hilti.com/c/CLS_MEA_TOOL_INSERT_7127/CLS_CONSTRUCTION_TOTAL_STATIONS_7127/r4728599}} automated Total Station or a MoCap system.
In fact, the focus of our dataset is to advance the state-of-the-art in SLAM in terms of accuracy with the final target of developing systems that are able to achieve millimeter level accuracy as required in construction environments.
\section{Hardware} \label{sec:Hardware}
Our sensor suite (the 'Phasma' stick, Fig.~\ref{figurelabel_phasma_stick}) consists of 3 categories of sensing modalities operating with different ranges and noise levels.  
These include:
\subsection{Passive Visual}
\begin{itemize}
\item[] \textbf{Alphasense by Sevensense\footnote{\url{https://www.sevensense.ai/product/alphasense-position}}} \\
The visual data is collected from an array of rigidly mounted 1.3 MP global shutter cameras. This module consists of 5 wide field-of-view cameras mounted to give an approximate 270 deg continuous field of view. Within this configuration, a stereo pair is present. Images are synchronously collected at 10 Hz. While the Alphasense can capture at a \rebuttal{higher} frame rate, doing so would have required using a lower resolution. Since this dataset aims for maximum accuracy, the higher resolution was chosen.

\end{itemize}

\begin{figure}
  \centering
  \includegraphics[width=\columnwidth]{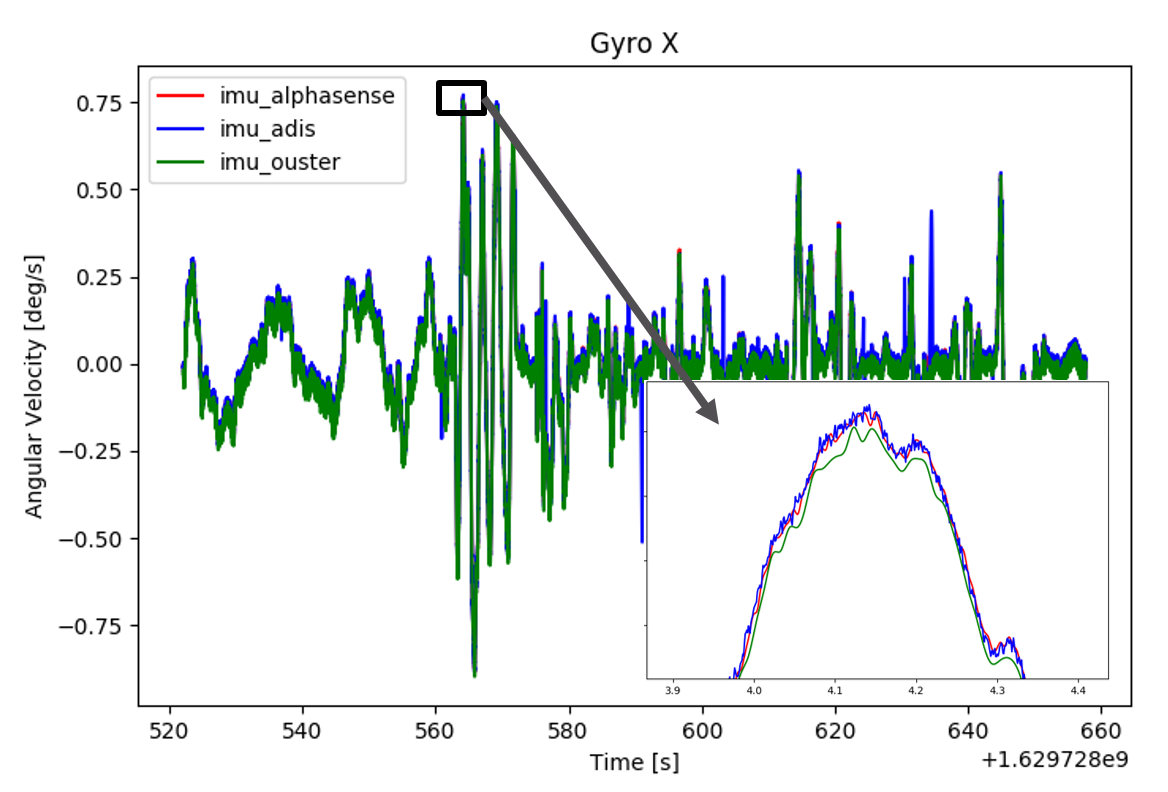}
  \caption{Verification of the time synchronization between the LiDAR and the camera module. The gyroscope data show a difference of \textless 1ms.}
  \label{figurelabel_imuSync}
\end{figure}

\subsection{Active Optical }
\begin{itemize}
\item[] \textbf{Ouster OS0-64\footnote{\url{https://ouster.com/products/os0-lidar-sensor/}}} \\
Long-range point cloud data is collected by the 360 deg scanning LiDAR sensor. This unit has a scan repetition of 10 Hz, and a point data rate of 1,300,000 points/second. Ranges are recorded from 0.3 to 50 m, with typical lowest noise returns greater than 1 m. Range accuracy is 1.5-5 cm.\\
\item[] \textbf{Livox MID70\footnote{\url{https://www.livoxtech.com/mid-70}}} \\
This unit is a LiDAR sensor with a 70 deg circular field of view and a non-repeating scan pattern. Point data rate is 100,000 points/second.  Ranges are recorded from 0.02 to 200 m, with typical returns between 1 and 50 m. Range accuracy is 2-5 cm.
\end{itemize}
\subsection{Inertial Sensors}
\begin{itemize}
\item[] \textbf{Analog Devices ADIS16445\footnote{\url{https://www.analog.com/en/products/adis16445.html}}} \\
This IMU is rigidly mounted to the AlphaSense module. It is a high-performing MEMS based sensor with relatively low noise and sensor bias drift rates. The data from this IMU is tightly timestamped to the AlphaSense timing system. Data is collected at 800 Hz.\\
\item[] \textbf{Bosch BMI085\footnote{\url{https://www.bosch-sensortec.com/products/motion-sensors/imus/bmi085/}}} \\
This IMU is embedded in the AlphaSense module. It provides a modest level of performance in terms of noise and bias stability. The data from this IMU is tightly timestamped to the AlphaSense timing system. Data is collected at 200 Hz.\\
\item[] \textbf{InvenSense ICM-20948\footnote{\url{https://invensense.tdk.com/products/motion-tracking/9-axis/icm-20948}}} \\
This IMU is embedded in the Ouster LiDAR. It provides a more modest level of performance than the ADIS16445 in terms of noise and bias stability. The data from this IMU is tightly timestamped to the Ouster timing system. Data is collected at 100 Hz.
\end{itemize}

\subsection{Ground Truth System}
For testing and validation, 2 systems are used to capture the ground truth:
\begin{itemize}
\item[] \textbf{Total Station} \\
A survey grade prism is attached to the Phasma stick, tracked by the Hilti PLT 300 automated Total Station. 
%Most datasets are collected in a "stop\ 'n\  go" fashion, where the total station precisely measures the prism during the 'stop' periods. Range measurements to the static prism have 3mm accuracy. Total station range and angle measurements are processed to generate XYZ position information. \\
\rebuttal{Ground truth data is collected in a "stop\ 'n\  go" fashion, where the total station precisely measures the prism during the 'stop' periods.
The stick is gravity aligned, using a mechanical system, before collecting each ground truth measurement.
The total station range and angle measurements are processed to generate XYZ position information.
In this situation, the range measurements of the static prism have accuracy of 3 mm.
}

\item[] \textbf{Optical Tracking} \\
Optical tracking targets are attached to the Phasma stick. When operated in a motion capture space, the multiple targets allow for the direct computation of the 6-DOF pose. Those ground truth data-points have a position accuracy of \textless 1mm and are collected at 200 Hz.
\end{itemize}

\subsection{DATA SYNCHRONIZATION AND LOGGING}

In a dynamic multi-sensor system, time synchronization between sensors is critical to make the best use of sensors fusion. Special care was given to synchronization of the Phasma stick to ensure maximum performance:\\
\begin{itemize}
\item[] \textbf{AlphaSense, Bosch IMU and ADIS IMU}\\
The AlphaSense manages time synchronization at the hardware level via an FPGA implementation. Camera times are computed to the mid-exposure pulse (MEP). Imu data is time tagged on arrival to the FPGA data bus. Overall time synchronization between the cameras and IMUs is \textless 1 ms.\\
\item[] \textbf{Ouster LiDAR and Invensense IMU}\\
The Ouster module includes an integrated IMU. The Ouster point data and IMU are hardware synchronized to the Ouster internal clock. Time synchronization between the two sensors is \textless 1ms.\\
\item[] \textbf{Cross Module Synchronization}\\
Synchronization between modules (AlphaSense, Ouster, Livox) is provided by the supported PTP network time protocol \cite{4579760}. Each module is attached via wired Ethernet cable to the data logging device, which hosts the PTP master clock. With this setup, the time alignment between the modules is observed to be \textless 1 ms, as shown in Fig. ~\ref{figurelabel_imuSync}. For verification, we adopted the approach from \cite{s21010068} and used optimization tools over the correlation signal of gyroscope data.\\
\end{itemize}
\noindent
Data logging occurs on a dedicated computer connected to the Phasma stick. The logging computer, with the Ubuntu 18.04 OS, runs the Robotic Operating System (ROS) during the data capture. Sensing modules are connected to the data logger, and data streams are directly recorded in ROS bag files.

\subsection{CALIBRATION}
Accurate intrinsic and extrinsic sensors calibration is critical to achieving the highest system performance. 
% Therefore, extensive calibration was undertaken to align the various sensors in our setup. 
Each respective manufacturer conducted intrinsic sensor calibration. 
We re-run the intrinsic calibration of the cameras using a standard checkerboard calibration.
For LiDARs, proprietary calibration models were used and corrections applied to the data at the time of capture.\\
The reference point of the body frame of the Phasma stick is defined at the AlphaSense Bosch IMU center.
All other sensors are transformed back to this point. 
\rebuttal{Extrinsic calibrations were determined from the CAD model of the Phasma design.}
%Spatial offsets were determined from the CAD model of the Phasma design.
\rebuttal{The calibration procedure proposed in~\cite{Furgale13iros} was used to refine the cameras-IMUs extrinsic calibrations.}
Calibration files, CAD Models, and sensor noise parameters are provided on the dataset website. 
\section{Dataset} \label{sec:Dataset}

\begin{figure}[h!t]
  \centering
  \includegraphics[width=\columnwidth]{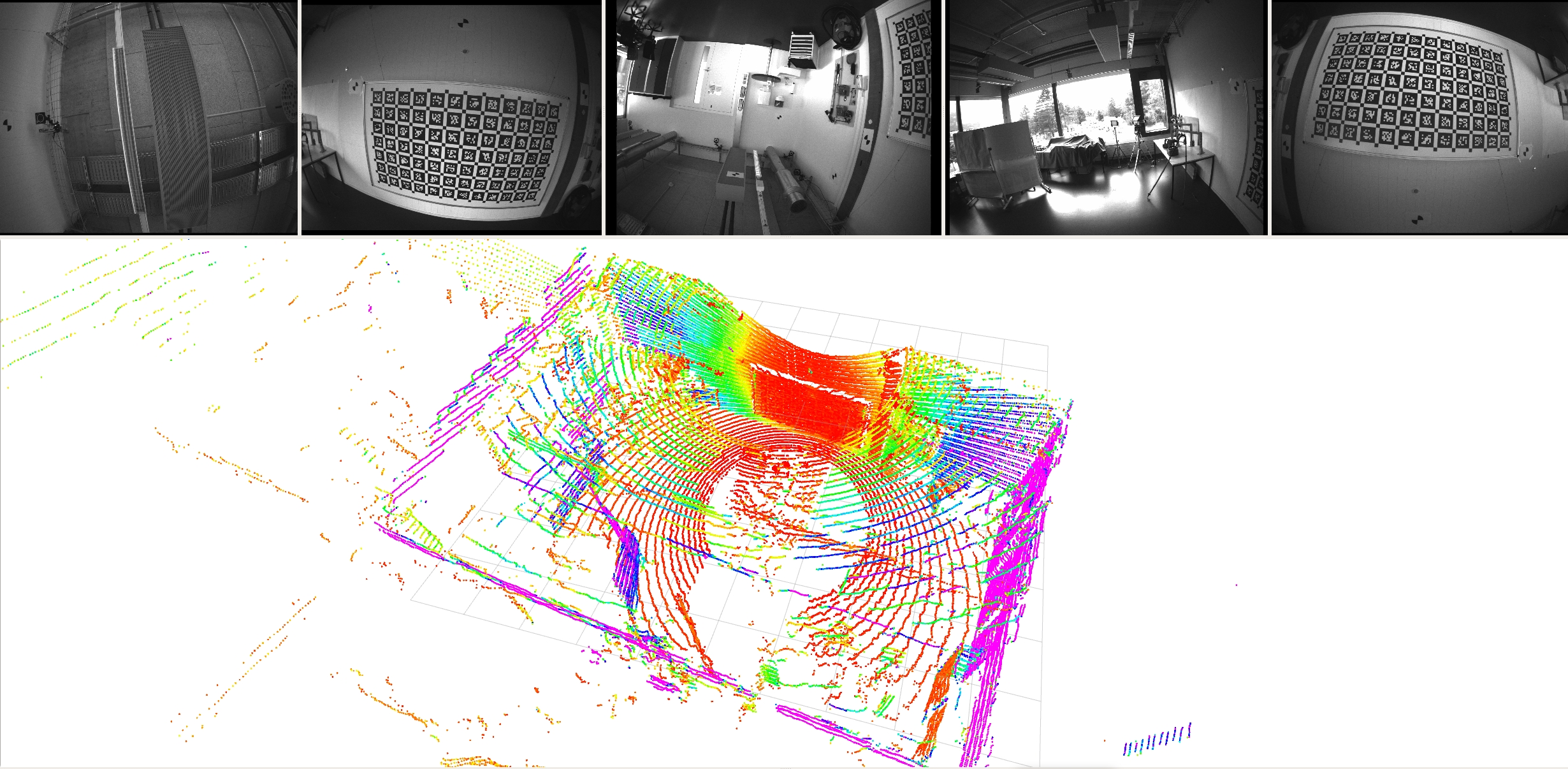}
  \caption{Example from the dataset 'Lab Survey 2': images from the 5 cameras and the Ouster OS0 point cloud}
  \label{figure:label_dataset_example}
\end{figure}

\begin{figure*}%
    \centering
    \subfloat[\centering]{{\includegraphics[width=.45\linewidth,valign=c]{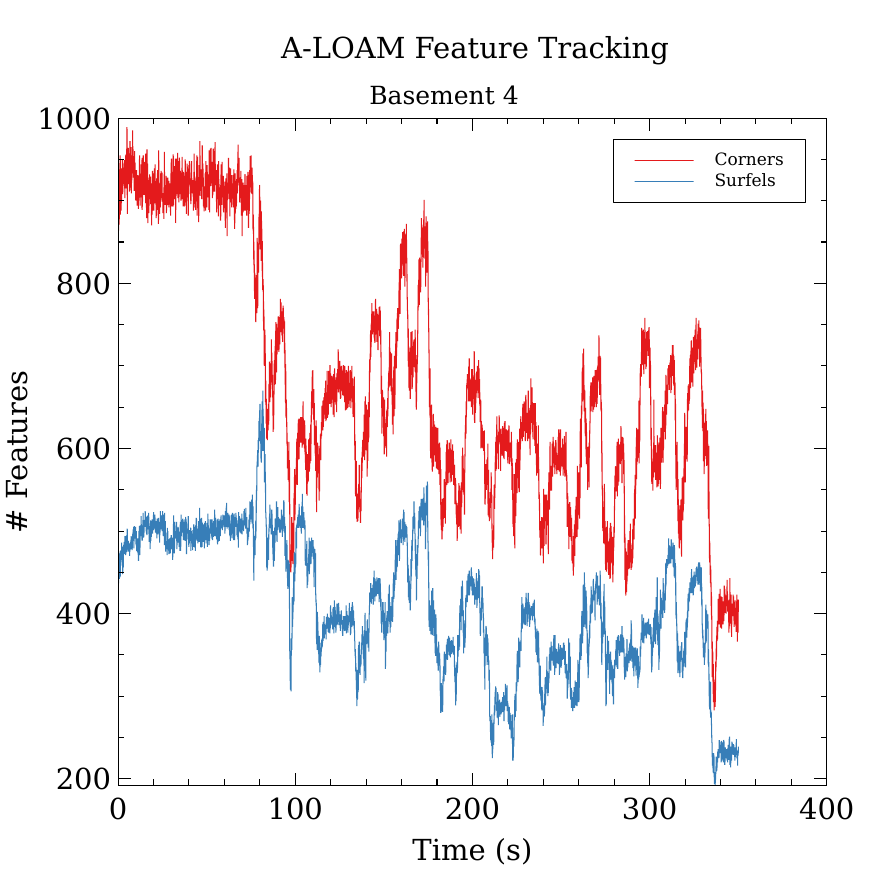} }}%
    \qquad
    \subfloat[\centering]{{\includegraphics[width=.45\linewidth,valign=c]{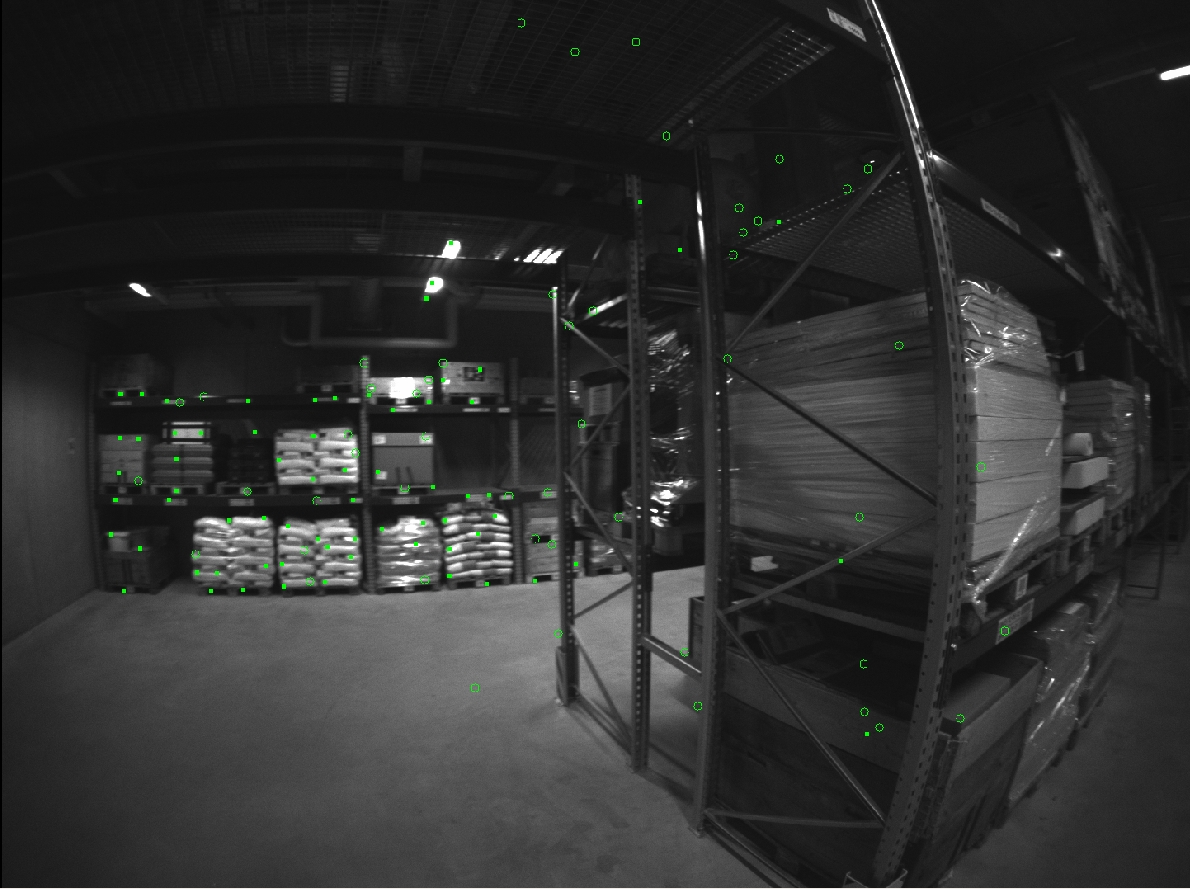} }}%
    \caption{(a) Features tracked in 'Basement 4' using the LiDAR-based SLAM algorithm A-LOAM. (b) the local environment during a low LiDAR feature event (time approx. 336 s). Close walls and structures reduce the LiDAR measurements, whereas vision based algorithms can still operate normally (the feature tracks of SVO2~\cite{7782863} are shown).}%
    \label{fig:feature_track}%
\end{figure*}

Data was collected under various conditions in indoor, outdoor, and mixed indoor-outdoor environments. The data shows practical challenges in different stages of constructions. Challenges include variable lighting, limited features and/or highly reflective and transparent surfaces.
\\ \\
Data descriptions (see Fig.~\ref{some_example_scenes}):
\begin{itemize}
\item[(a)] \textbf{Basement}\\
Data was collected in a windowless room (approx. 20x40 m). No natural light, mixed illumination brightness. Concrete space with building infrastructure. Basement 1 is a short and easy path. In Basement 3 and Basement 4, we mounted the sensor platform on a moving base instead of operating it handheld. Basement 3 and 4 also allow to exploit loop closure capabilities of the SLAM systems.
\item[(b)] \textbf{Campus}\\
Data was collected outdoors in a courtyard setting (approx. 40x60 m). Good natural lighting with high illumination. Mixed features with building structure and natural flora.
\item[(c)] \textbf{Construction Site}\\
Mostly outdoors with some covered areas (approx. 40x80 m). Strong natural light with high illumination. Unfinished natural surfaces with limited features above the ground plane.
\item[(d)] \textbf{IC Office}\\
Indoor space with many windows and reflective surfaces (approx. 10x70 m). Mix of natural and artificial light. Strong illumination at the windows, modest illumination indoors.
\item[(e)] \textbf{Lab}\\
Indoor space dominated by large windows (approx. 10x10 m). Strong natural light and reflective surfaces. Optitrack 6-DOF ground truth.
\item[(f)] \textbf{Office Mitte}\\
Indoor space in finished office building (approx. 30x50 m). Mix of natural and artificial light. Lots of building structure.
\item[(g)] \textbf{Parking}\\
Mix of indoor and outdoor space (approx. 100x100 m). Parking garage from the top floor to the bottom floor. Lighting varies from extreme bright to modest darkness. Ground plane structure on top floor and plenty of building structure on the lower floor.
\item[(h)] \textbf{RPG Tracking Area}\\
Indoor test facility (approx. 30x30 m). Mostly artificial light with some natural. Single large room with random motion path throughout. MoCap 6-DOF ground truth.
\end{itemize}

\subsection{Dataset Format}

Datasets are stored in binary format (rosbags) which contain images and IMU and LiDAR data. For the data from the Livox a custom message was chosen as it contains additional timing information compared to the standard \texttt{ROS PointCloud2} message. 
Fig.~\ref{figure:label_dataset_example} shows an example of the camera and LiDAR data from the Lab Survey 2 dataset. Ground truth data is given in a separate file for each dataset, with the filename indicating the reference source (e.g. \texttt{Construction\_Site\_prism.txt} means the ground truth is in the \textit{prism} frame). All topics in the bag are listed in Table \ref{table:rostopics}.
\begin{table*}[h!]
\caption{Message topics and types of the rosbags}
\centering
\begin{tabular}{lll}
\textbf{Topic}                                  & \textbf{Type}                         & \textbf{Description}          \\
\texttt{/alphasense/cam0/image\_raw}            & \texttt{sensor\_msgs/Image}           & front facing camera 1 (right)        \\
\texttt{/alphasense/cam1/image\_raw}            & \texttt{sensor\_msgs/Image}           & front facing camera 2 (left)        \\
\texttt{/alphasense/cam2/image\_raw}            & \texttt{sensor\_msgs/Image}           & upward facing camera          \\
\texttt{/alphasense/cam3/image\_raw}            & \texttt{sensor\_msgs/Image}           & right facing camera            \\
\texttt{/alphasense/cam4/image\_raw}            & \texttt{sensor\_msgs/Image}           & left facing camera           \\
\texttt{/alphasense/imu         }               & \texttt{sensor\_msgs/Imu  }           & Bosch IMU, 200Hz              \\
\texttt{/alphasense/imu\_adis    }              & \texttt{sensor\_msgs/Imu  }           & \rebuttal{ADIS16445}, 800Hz              \\
\texttt{/livox/lidar            }               & \texttt{livox\_ros\_driver/CustomMsg} & Livox MID70                   \\
\texttt{/os\_cloud\_node/imu}                   & \texttt{sensor\_msgs/Imu}             & Ivensense, 100Hz              \\
\texttt{/os\_cloud\_node/points }               & \texttt{sensor\_msgs/PointCloud2}     & Ouster OS0-64                              \\
\texttt{tf\_static }                            & \texttt{tf2\_msgs/TFMessage}          & all transforms between frames
\end{tabular}
\label{table:rostopics}
\end{table*}
\subsection{Challenges}

This section includes an example of the challenges contained in our dataset and highlights the need of multiple sensors fusion.
In the sequence \textit{Basement 4}, the number feature tracks by the LiDAR odometry algorithm, A-LOAM~\cite{zhang2014loam}, remains high (around 1000 per frame) while the Phasma device is in the centre of the room, but it decreases significantly when the device approaches a wall or a close overhang, see in Fig. \ref{fig:feature_track} left.
In this case, the accuracy of the estimated trajectory by A-LOAM is negatively affected. 
However, the decrease of the LiDAR feature tracks can be compensated by the increase of the visual features, see Fig.~\ref{fig:feature_track} right. 
In this case, the best performance can be achieved by an algorithm that fuses camera and LiDAR measurements.

\section{The Hilti SLAM Challenge} \label{sec:Challenge}

\begin{table*}[h!t]
\vspace*{0.5cm} 
\caption{Results of the Hilti SLAM Challenge. The reports including the description of each submission can be found on the dataset website.}
\centering
\begin{tabular}{l|l|l|l|c|c}
\textbf{Team} & \textbf{Method}& \textbf{Description} &\textbf{Sensors} &\textbf{RMSE {[}m{]}} & \textbf{Score} \\ \hline
Megvii3D & based on~\cite{fastlio} & IEFK based & LiDAR + imu & 0,0774 & 461 \\
Bosch Research & closed-source & graph optimization, Manhattan world & LiDAR + imu & 0,0957 & 457 \\
Vision \& Robotics GmbH & based on~\cite{Neuhaus2018MC2SLAMRI} & tightly coupled MHE, loop closure, BA & LiDAR + imu & 0,1010 & 406 \\
GeoSLAM &  closed-source & sliding window, loop closure, BA & LiDAR + imu & 0,2602 & 389 \\
Oxford   Robotics Institute & VILENS~\cite{wisth2021vilens} & tightly coupled, based on factor graphs & camera + LiDAR + imu  & 0,1546 & 378 \\
Nanyang Technological University  & VIRAL~\cite{nguyen2021viral} & sliding window, BA &  camera + LiDAR + imu& 0,2024 & 346 \\
CMU Doom & closed-source & sliding window + loop closure & camera + LiDAR + imu & 0,3592 & 333 \\
ETH Zürich & Maplab~\cite{2018maplab} & tightly coupled, graph based  & camera + LiDAR + imu &  0,7467 & 288 \\
NPM3D Team, MINES ParisTech & CT-ICP~\cite{dellenbach2021cticp} & scan-to-map, no loop closure, no ba & LiDAR &3,6302 & 273 \\
UC San Diego & closed-source & scan-to-map, imu for undistortion  & LiDAR + imu & 0,0838 & 258 \\
C.F Rubio et.al &  based on~\cite{aloam} & optimization based & LiDAR + imu & 8,6470 & 228 \\
IVISO & closed-source & graph based approach, no loop closing & camera + imu & 0,6135 & 219 \\
Spectacular AI & HybVIO~\cite{seiskari2021hybvio} & MSCKF~\cite{4209642} based, no global ba & camera + imu &0,6758 & 216
\end{tabular}
\label{table:challenge_results}
\end{table*}

The proposed dataset was used in the \textit{Hilti SLAM challenge}, which first edition took place at the 2021 IEEE/RSJ IROS conference. 
Academic and industrial groups submitted the solutions of their SLAM algorithms on all the sequences of the dataset.
Participants had access to the ground truth of half of the sequences. 
The other half was used for evaluation (similarly to what is done in KITTI~\cite{Geiger2012CVPR}).
An accuracy-dependent score was computed for ranking teams: after SO3 alignment~\cite{8593941} of the estimated trajectory with the ground truth, every single point scored between 10 and 0 points: 10 points for errors below 1 cm, 6 points between 1 cm and 10 cm and 3 points for errors up to 1 m. This format was chosen, instead of computing the ranking based on the \rebuttal{RMSE}, because in this way we could also consider non-complete or missing trajectories, which would have distorted the ranking otherwise. \rebuttal{The choice of the thresholds was based our use-case, which requires sub-cm accuracy.} In total, 27 teams participated in the challenge, with 7 commercial companies among them, see \rebuttal{Table \ref{table:challenge_results}}. 
The first four places have been taken by commercial algorithms, which all focused on LiDAR-IMU odometry, showing  the maturity and robustness of these approaches. The best team, Megvii, used a variant of FAST-LIO2 \cite{fastlio} and achieved an average error of 9.3 cm on all sequences. Megvii was one of the few teams that merged the Ouster and the Livox LiDAR data, which, together with using all LiDAR points for state estimation, gave them a significant advantage. The best algorithm that fuses vision with LiDAR and imu ranked 5th, VILENS~\cite{wisth2021vilens} by the Oxford Robotics Institute. The best vision-only solution ranked 12th, with the majority of errors larger than 50 cm. 
The results show that, unsurprisingly, commercial algorithms outperform academic algorithms. The exact gap, however, was not clear prior to this challenge. The results also show that there is still room for improvement since the winning team did not fuse camera data. Table~\ref{table:challenge_results} shows an overview of the teams and their approaches (teams who decided to stay anonymously are not shown in the table).
\section{Known Issues} \label{sec:KnownIssues}
Despite careful design and execution of the data collection experiments, we are aware of some issues which pose additional challenges for SLAM algorithms and limit the achievable. These include:
\begin{itemize}
\item  Clock drift and offset: The clock from MoCap and the data logging computer are not hardware-synchronized. We used Ethernet connection and a time-of-arrival time stamping to keep the offset to a minimum, however we observed a difference of around 1-3ms in the two clocks.
\item Some frames in the LiDAR, camera and IMU data were dropped due to high load on the controller. 
\end{itemize}
\section{Conclusion} \label{sec:Conclusion}
In this paper, we described a new public dataset captured with a redundant multi-sensor platform containing visual, inertial, and LiDAR data. 
The dataset includes a series of real-world scenarios collected with state-of-the-art sensing technologies and high-quality time-synchronization. 
Our goal is to foster research in SLAM to advance the current state-of-the-art and make SLAM systems ready to be deployed in real-world applications with demanding requirements in terms of accuracy and robustness, such as in construction robotics.
The dataset was used in the \textit{Hilti SLAM Challenge}.
The results of the challenge show a picture of the capabilities of the current SLAM algorithms, from both academia and industry, and potential improvements, such as multi-sensors fusion.
\section{Acknowledgments } \label{sec:Acknowledgments }
We thank Danwei Wang from Nanyang Technological University, Singapore, and Yufeng Yue from Beijing Institute of Technology, China, for hosting the challenge in their IROS 2021 workshop.

%%%%%%%%%%%%%%%%%%%%%%%%%%%%%%%%%%%%%%%%%%%%%%%%%%%%%%%%%%%%%%%%%%%%%%%%%%%%%%%%

\printbibliography %Prints bibliography

\end{document}